\documentclass[12pt,a4paper]{article}
\usepackage[a4paper,left=2.5cm,right=2.5cm,top=2.5cm,bottom=2.5cm]{geometry}

\usepackage{url}

\usepackage{float}

\usepackage{algorithm}	
\usepackage[noend]{algpseudocode}
\usepackage{amsmath}
\usepackage{amssymb}
\usepackage[title]{appendix}
\usepackage{adjustbox,lipsum}
\usepackage{threeparttable}
\usepackage{graphicx}

\algdef{SE}[SUBALG]{Indent}{EndIndent}{}{\algorithmicend\ }%
\algtext*{Indent}
\algtext*{EndIndent}

\begin{document}
\date{}
\title{Distance approximation using Isolation Forests}
\author{David Cortes}
\maketitle

\begin{abstract}
This work briefly explores the possibility of approximating spatial distance (alternatively, similarity) between data points using the Isolation Forest method envisioned for outlier detection (\cite{iso1}). The logic is similar to that of isolation: the more similar or closer two points are, the more random splits it will take to separate them. The separation depth between two points can be standardized in the same way as the isolation depth, transforming it into a distance metric that is limited in range, centered, and in compliance with the axioms of distance. This metric presents some desirable properties such as being invariant to the scales of variables or being able to account for non-linear relationships between variables, which other metrics such as Euclidean or Mahalanobis distance do not. Extensions to the Isolation Forest method are also proposed for handling categorical variables and missing values, resulting in a more generalizable and robust metric.
\end{abstract}

\section{Introduction}

This work explores the idea of using Isolation Forests (\cite{iso1}, \cite{iso2}) and variations thereof (\cite{extiso}, \cite{sciforest})  for estimating how similar/closer or dissimilar/further two points are in an arbitrary feature space ${\rm I\!R}^n$, based on an observed sample of data to which an Isolation Forest model is fit, and from which the distributions and relationships between different variables/dimensions are implicitly incorporated into this distance/similarity metric.

The premise is simple: if a set of data points is split into two branches recursively multiple times by choosing some variable and split point in that variable uniformly at random, the points that are more distant will on average be separated (put into different tree branches) with fewer splits (closer to the root node), while points that are more similar will require more splits to become separated.

The procedure is less efficient than simpler calculations such as Euclidean distance (${\lVert} x_1 - x_2 {\rVert}_2$), but offers some advantages: this distance/similarity is invariant to the scale of each variable, having non-normal distributions does not present any issues, and potential relationships between variables in the distribution are taken into account, even if these relationships are not linear. Additionally, some small modifications to the Isolation Forest algorithm allow incorporation of categorical variables and handling of missing values in the procedure.

\section{Isolation Forests}

Isolation Forest (a.k.a. iForest, \cite{iso1}, \cite{iso2}) is an algorithm devised for outlier or anomaly detection based on the concept of isolation: if a set of data points is split according to some random variable by finding a split point at random within the range in the data, assigning all points that are less or equal than this threshold to one branch and the rest to the other, and this process is continued recursively on each branch, then outlier points will become isolated (put alone) in one branch quicker (with fewer splits, closer to the root node of the tree) than non-outlier points.

The idea can be extended to non-random splits based on the standard deviations of the variable being split that are obtained at each branch (\cite{sciforest}), and to splitting hyperplanes (\cite{extiso}, \cite{sciforest}), which as shown in \cite{extiso}, can help to remove some biases that are introduced by the single-variable splitting process.
As outliers can only be considered to be so if their average isolation depth is less than expected for a random data point, this procedure can be terminated before isolating every single point by stopping the process once it reaches the depth that a balanced binary tree would have, and the remainder isolation depth for non-isolated points approximated by adding to the terminal depth the expected value of this depth for each point if the process were continued with uniformly-random data and uniformly-random splits on the number of points that remain on that node.

The average isolation depth obtained for a given point can be converted into a standardized outlier metric according to how it differs from the expected isolation depth for a random data point, which is given by $2(H_n - 1)$, where $H_n$ is the nth harmonic number - see \cite{iso2} and \cite{depth} for details.

\section{Separation depth and distance}

If considering two random points in a subset, one can also think of separation instead of isolation as the  binary trees are grown: if the points are split (assigned to different branches from a binary tree node) according to being smaller or greater than a random value within the range of some variable in the feature space, then the closer two points are in that dimension, the higher the probability that they will end up in the same branch if the split point is chosen uniformly at random, due to the fact that, the closer they are, the larger the number of possible split points under which they end up together, and if some variable underwent a linear or affine transformation $\widetilde{\mathbf{x}} = a \mathbf{x} + b$, each possible split on the points will still have the same probability as before, as this only depends on their relative position within the range of the variable.

If the process is repeated further, choosing a variable and split point uniformly at random in each branch that was obtained in the previous split, then again closer points in the new variable have a higher chance of ending up in the same branch, but this time it is conditioned on already not having been separated in the previous split. In this regard, non-random splits that aim at finding the point that minimizes the standard deviations of the variable in the obtained branches could also do a better job at making clustered points appear even more similar, due to the fact that splits will tend to separate clusters first (see \cite{sciforest} and \cite{c45}).

\begin{figure}[tph!]
\centerline{\includegraphics[totalheight=5cm]{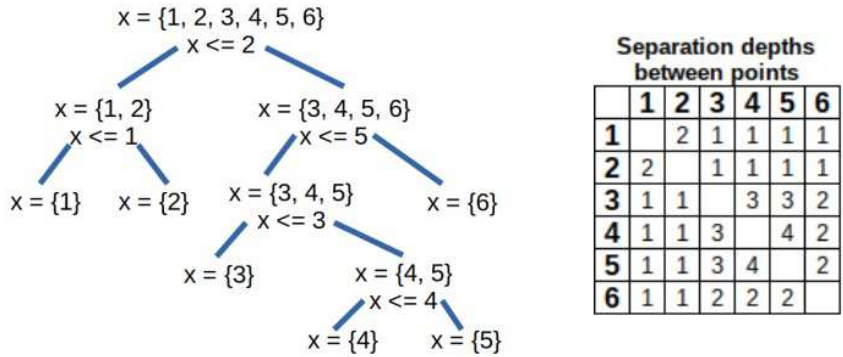}}
    \caption{Example random tree on 1-d data points}
    \label{fig:verticalcell}
\end{figure}

If this procedure is repeated indefinitely from the beginning until each point becomes isolated, then the \textbf{average} separation depth between any two points across these random trees will be greater iff the points are closer (with closeness influenced by the data distribution), and having multiple trees will remove the large expected variability introduced by having to start by separating points according to one variable (that is, a large number of pairs is expected to become separated with the first split, which is not what happens with isolation as most points don't end up isolated after the first split).

Just like with isolation, it's possible to calculate the expected separation depth between two random points if a set of points is split by a random tree procedure which would assign from the remaining $m$ points a random number $\sim \text{Uniform}(1, m-1)$ of them to go to one branch and the remainder to the other.
The expected separation depth under a randomly-built tree like this with the same number of terminal nodes as points in the data or subset in a terminal node, can be calculated recursively by considering that, if two points are not separated right after a split, they will go together into yet another node, but of smaller remainder size, in which the procedure will be repeated again - see \cite{sep} for details. The formula is given by:
$$
\mathbb{E}[s_n] = 1 + \frac{1}{n-1}\sum_{i=1}^{n-1} ( 
	\frac{{i \choose 2}}{{n \choose 2}} \mathbb{E}[s_i]	
	+
	\frac{{n-i \choose 2}}{{n \choose 2}} \mathbb{E}[s_{n-i}]	
  )
$$
With $\mathbb{E}[s_1]=0$ (single point is already isolated) and $\mathbb{E}[s_2]=1$ (two points always become separated in one split).

It can be more efficiently calculated by a recursion as follows:
$$
\mathbb{E}[s_n] = \frac{-n \mathbb{E}[s_{n-1}] + 3 n - 4}{n (n - 1)}
$$

As the sample size grows to infinity, the expected separation depth can be more easily approximated: this scenario with infinite discrete choices is equivalent to a scenario with a continuous number line, for which at each split, two points plus a split threshold are drawn according to a random uniform distribution, and will be separated if the threshold lies inbetween the two points - the probability of this happening in a uniform distribution is $\frac{1}{3}$ regardless of the range, and if they are not separated in that split, then the process starts again with a still-infinite sample, which will give the same probability of $\frac{2}{3}$ of not becoming separated, and thus the expected separation depth is $\mathbb{E}[s_{\infty}] = 1 + \lim_{x \to \infty}\sum_{i=1}^{x} (1 - \frac{1}{3})^i = 3$.

The fact that this number is constant for indefinitely large sample sizes comes in handy, as then one can assume that this is what it takes to separate two points in an arbitrary data sample as will be explained next.

For determining isolation depth, when a tree node reaches the height limit with multiple points (as opposed to a single isolated point) or contains a set of points in which no further split is possible, the expected isolation depth for that remainder in \cite{iso1} is approximated according to how many points ended up there, and this number is taken again at prediction time for new data points if they reach that terminal node in the tree. For separation, this is problematic as at the moment of determining the distances between points in a terminal node, the expected remaining separation depth will depend on the number of points that end up in that terminal node, making the distances between two points be affected by the presence of a third point when the trees are grown or if this expectation were to be determined in the same way as isolation for new points. Fortunately, there would not be such a dependence on a third point when it comes the time to use the already-fitted trees to estimate the distance between two new points if it can be assumed that at the end of each terminal node will lie an infinite sample of more data points drawn according to the same data distribution, because the estimated separation depth for an infinite sample is always $3$ so if the separation depth is calculated for a new sample of points, once two of them reach the same terminal node in a tree, their expected separation depth will be the same as if there were yet more points reaching that same terminal node.

Just like for the outlier score proposed in \cite{iso1}, the expectation in a randomly-built tree can also be used to produce a standardized metric, by comparing the obtained average separation depth between points against the expected separation depth in a randomly-built tree through a simple transformation such as $f(s) = 2^{-\frac{s-1}{\mathbb{E}[s]-1}} = 2^{-\frac{s-1}{2}}$ - subtracting the minimum separation from both numbers so as to make the metric be able to reach its maximum. This standardized metric, which measures dissimilarity due to the negative sign, presents some nice properties: on average, points should have a dissimilarity between them of about $0.5$ ($= 2^{-1}$), with points that are more similar than average having values closer to zero, and points that are more dissimilar than average having values closer to $1$.

This dissimilarity (from here on, distance) can be shown to be a proper metric distance under some extra assumptions:
\begin{itemize}
	\item It is bounded between zero and one ($\lim_{x \to \infty} 2^{-\frac{x}{2}} = 0$, $\lim_{x \to 0} 2^{-\frac{x}{2}} = 1$), thus $d(\mathbf{x}_1, \mathbf{x}_2) \geq 0 \:\: \forall \:\: \mathbf{x}_1, \mathbf{x}_2 \in \rm I\!R^n$.
	\item If it is assumed that a single point is indivisible and thus it's average separation depth infinite, then $d(\mathbf{x}_1, \mathbf{x}_1) = \lim_{x \to \infty} 2^{-\frac{x}{2}} = 0$.
	\item Since (i) there is only one possible path between any two points in a given tree, (ii) each pair of points requires at least 1 split to be separated, and (iii) a point (a) cannot be separated from point (b) further down the tree than it is separated from point (c) if points (b) and (c) have already been separated earlier; it means $s(a, c) < s(a, b) + s(b, c)$ in every tree, which implies $d(\mathbf{x}_1, \mathbf{x}_3) \leq d(\mathbf{x}_1, \mathbf{x}_2) + d(\mathbf{x}_2, \mathbf{x}_3)$.
\end{itemize}

This analysis assumes that all points are unique, and the expected separation calculation would not work if duplicated points are passed down the tree as then they would be considered to have some positive distance even though they are the same. This can however be taken into account at the moment of growing the trees by assigning them a higher sampling weight, and duplicates can be filtered out before being passed through already-fitted trees.

\section{Categorical variables and missing values}

It's possible to think of some simple extensions to the original Isolation Forest model for handling categorical variables as follows: a random subset of the present categories is assigned to one tree branch, while the rest are assigned to the other tree branch, and when new points are passed down the tree, if they have a category that was not present in the original points from which the split was determined, they are divided heuristically, either by assigning them to the branch that had the fewer points, or by assigning them to both branches but with a weight given by the proportion of points before the split that were assigned to each branch, and the results later combined according to these weights. This same trick can be used for handling missing data, oftentimes providing better results than a-priori imputation (see \cite{nas}). In the extended model (\cite{extiso}, \cite{sciforest}), which produces splits by more than one variable at a time, missing data and new categories can alternatively be imputed with the median of the sample or sub-sample from which the split was determined (i.e. only the points that reach that current level). In the case of categorical variables, each category would have its own coefficient to add to the linear combination, and the resulting numeric transformation under these coefficients will have some median in the original sample that can be used as imputation value.

The full procedure is described below:

\begin{algorithm}[H]
\caption{iForestEnhanced}\label{iForestEnhanced}
\hspace*{\algorithmicindent} \textbf{Inputs} $\mathbf{X}$ (input data with dimensionality $p$), $t$ (number of trees), $n$ (sub-sample size), $m$ (number of splitting dimensions), $d$ (max depth) \\
\hspace*{\algorithmicindent} \textbf{Output} Isolation Forest model consisting of $t$ trees
\begin{algorithmic}[1]
\State Start with empty set of trees $F = \emptyset$
\For {$1..t$}
	\State Take subsample $\mathbf{X}_t$ consisting of $n$ points from $\mathbf{X}$ selected randomly.
	\If {$m = 1$}
		\State Add single-variable tree: $F := F \cup \text{iTreeEnh}(\mathbf{X}_t, d, 0, \mathbf{1}^n)$ 
	\Else
		\State Add extended tree: $F := F \cup \text{iTreeExtEnh}(\mathbf{X}_t, d, 0, m)$
	\EndIf
\EndFor
\Return $F$
\end{algorithmic}
\end{algorithm}

\begin{algorithm}[H]
\caption{iTreeEnh}\label{iTreeEnh}
\hspace*{\algorithmicindent} \textbf{Inputs} $\mathbf{X}$ (input data points), $d$ (max depth), $h$ (current depth), $\mathbf{w}$ (weight of each point in $\mathbf{X}$) \\
\hspace*{\algorithmicindent} \textbf{Output} Tree node with left branch $T_l$, right branch $T_r$, proportion left $b_l$, chosen variable $y$, present categories $\textsc{c}$, and either split point $z$ or split subset $S$
\begin{algorithmic}[1]
\If {$|\{\mathbf{x} \in \mathbf{X}\}| = 1$ or $h = d$}
	\State Terminate procedure (return empty output $\emptyset$)
\Else
	\State Choose variable $y$ at random from $1..p$ such that $\mathbf{X}_{:,y}$ has at least 2 different values (if not possible, then terminate)
	\If {$y$ is numeric}
		\State Choose a random point $z \sim \text{Uniform}(\min{\mathbf{X}_{:,y}}, \max{\mathbf{X}_{:,y}})$
		\State Determine subsets $\mathbf{X}_l = \{ \mathbf{x}_i \in \mathbf{X} \:|\: x_{i,y} \leq z \}$, $\mathbf{X}_r = \{ \mathbf{x}_i \in \mathbf{X} \:|\: x_{i,y} > z \}$, $\forall x_{i,y} \:\:\text{known}$
		\State Set empty present categories $\textsc{c} = \emptyset$
	\EndIf
	\If {$y$ is categorical}
		\State Determine present categories $\textsc{c} = \{i \:\: | \:\: \exists i \in \mathbf{X}_{:,y} \}$
		\State Choose a random subset $S$ of categories from all possible subsets of $\textsc{c}$
		\State Determine subsets $\mathbf{X}_l = \{ \mathbf{x}_i \in \mathbf{X} \:|\: x_{i,y} \in S \}$, $\mathbf{X}_r = \{ \mathbf{x}_i \in \mathbf{X} | x_{i,y} \notin S \}$, $\forall x_{i,y} \:\:\text{known}$
	\EndIf
	\State Determine proportion assigned to first branch $b_l = \frac{\sum_{i | \mathbf{x}_i \in \mathbf{X}_l} w_i }{  \sum_{i | \mathbf{x}_i \in \mathbf{X}_l} w_i + \sum_{j | \mathbf{x}_j \in \mathbf{X}_r} w_j  }$
	\State Update weights $\mathbf{w}_l = \{ w_i \:|\: \mathbf{x}_i \in \mathbf{X}_l \} \cup \{ b_l w_i \:|\: x_{i,y} \:\text{is unknown} \}$
	\State Update weights $\mathbf{w}_r = \{ w_j \:|\: \mathbf{x}_j \in \mathbf{X}_r \} \cup \{ (1-b_l) w_j \:|\: x_{j,y} \:\text{is unknown} \}$
	\State Divide missing values s.t. $\mathbf{X}_l = \mathbf{X}_l \cup \{ \mathbf{x}_i \in \mathbf{X} \:|\: x_{i,y} \:\text{is unknown} \}$, $\mathbf{X}_r = \mathbf{X}_r \cup \{ \mathbf{x}_j \in \mathbf{X} \:|\: x_{j,y} \:\text{is unknown} \}$
	\State \Return tree node with left branch $T_l = \text{iTreeEnh}(\mathbf{X}_l, d, h + 1, \mathbf{w}_l)$, right branch $T_r = \text{iTreeEnh}(\mathbf{X}_r, d, h + 1, \mathbf{w}_r)$, left branch proportion $b_l$, chosen variable $y$, present categories $\textsc{c}$, and either split point $z$ or split subset $S$
\EndIf
\end{algorithmic}
\end{algorithm}

\begin{algorithm}[H]
\caption{iTreeExtEnh}\label{iTreeExtEnh}
\hspace*{\algorithmicindent} \textbf{Inputs} $\mathbf{X}$ (input data points), $d$ (max depth), $h$ (current depth), $m$ (number of splitting dimensions) \\
\hspace*{\algorithmicindent} \textbf{Output} Tree node with left branch $T_l$, right branch $T_r$, subset of variables $\textsc{u}$, chosen numeric coefficients $\textsc{z}$, categorical coefficients $\{ S_c \}$, imputation values $\textsc{r}$, split point $q$
\begin{algorithmic}[1]
\If {$|\{\mathbf{x} \in \mathbf{X}\}| = 1$ or $h = d$ or $w \leq 1$}
	\State Terminate procedure (return empty output $\emptyset$)
\Else
	\State Initialize linear combination $\mathbf{y} := \mathbf{0}$ for each point in $\mathbf{X}$
	\State Initialize empty set of numeric coefficients $z := \emptyset$ and categorical coefficients $S := \emptyset$
	\State Choose a subset $\textsc{u}$ of $m$ variables at random at random from $1..p$ such that $\mathbf{X}_{:,y}, y \in u_m$ has at least 2 different values (fewer than $m$ if not possible, terminate if none has at least 2 different values)
	\For {each numeric variable $v \in \textsc{u}$}
		\State Draw a random coefficient $z_v \sim \text{Normal}(0, 1)$
		\State Standardize coefficient as $z_v := \frac{z_v}{\sigma_{\mathbf{X}_{:,v}}}$
		\State Update $y_i := y_i + z_v x_{i,v} \:\: \forall \:\: x_{i,v} \:\: \text{known}$
		\State Add coefficient to set $\textsc{z} := \textsc{z} \cup z_v$
		\State Determine imputation value as $r_v = \text{Median}\{z_v x_{i,v} \:\forall\: x_{i,v} \:\text{known} \}$
		\State Update $y_i := y_i + r_v  \:\: \forall \:\: x_{i,v} \:\: \text{unknown}$
		\State Add imputation value to the set $\textsc{r} := \textsc{r} \cup r_v$
	\EndFor
	\For {each categorical variable $c \in \textsc{u}$}
		\State For each category, choose a random coefficient $s_i^c \sim \text{Normal}(0, 1) \:\: \forall \:\: i \:\:\:\: s.t. \:\: \exists i \in \mathbf{X}_{:,c}$
		\State Update $y_i := y_i + s_{x_{i,c}}^c \:\: \forall \: x_{i,c} \:\: \text{known}$
		\State Add set of coefficients to set $S := S \cup \mathbf{s}^c$
		\State Determine imputation value as $r_c = \text{Median}\{s_{x_{i,c}}^c\}$
		\State Update $y_i := y_i + r_c  \:\: \forall \:\: x_{i,c} \:\: \text{unknown}$
		\State Add imputation value to the set $\textsc{r} := \textsc{r} \cup r_c$
	\EndFor
	\If {$\min{\mathbf{y}} = \max{\mathbf{y}}$}
		\State Terminate procedure (return empty output $\emptyset$)
	\EndIf
	\State Choose a random split point $q \sim \text{Uniform}(\min{\mathbf{y}}, \max{\mathbf{y}})$
	\State Determine subsets $\mathbf{X}_l = \{ {\mathbf{x}_i \in \mathbf{X} \:|\: y_i \leq q} \}$, $\mathbf{X}_r = \{ {\mathbf{x}_j \in \mathbf{X} \:|\: y_j > q} \}$.
	\State \Return tree node with left branch $T_l = \text{iTreeExtEnh}(\mathbf{X}_l, d, h + 1, m)$, right branch $T_r = \text{iTreeExtEnh}(\mathbf{X}_r, d, h + 1, m)$, subset of variables $\textsc{u}$, numeric coefficients $\textsc{z}$, categorical coefficients $S$, imputation values $\textsc{R}$, split point $q$
\EndIf
\end{algorithmic}
\end{algorithm}

\begin{algorithm}[H]
\caption{SepDepth}\label{SepDepth}
\hspace*{\algorithmicindent} \textbf{Inputs} $\mathbf{X}$ (input data consisting of $n$ points), $t$ (number of trees), $F$ (isolation forest) \\
\hspace*{\algorithmicindent} \textbf{Output} Distance matrix $\mathbf{D}^{n \times n}$
\begin{algorithmic}[1]
\State Initialize pairwise sums of separation depths as $\mathbf{D} = \mathbf{0}^{n \times n}$
\If {the trees are single-variable}
	\State Initialize weights $\mathbf{w} = \mathbf{1}^n$
\EndIf
\For {each tree $T \in F$}
	\If {the trees are single-variable}
		\State Update $\mathbf{D} := \text{TraverseTree}(T, \mathbf{X}, \mathbf{D}, \mathbf{w})$ 
	\Else
		\State Update $\mathbf{D} := \text{TraverseExtTree}(T, \mathbf{X}, \mathbf{D})$ 
	\EndIf
\EndFor
\State \Return $2^{-\frac{\frac{\mathbf{D}}{|F|}-1}{2}}$
\end{algorithmic}
\end{algorithm}

\begin{algorithm}[H]
\caption{TraverseTree}\label{TraverseTree}
\hspace*{\algorithmicindent} \textbf{Inputs} $T$ (node of an iTreeEnh), $\mathbf{X}$ (input data), $\mathbf{D}$ (current sum of separation depths), $\mathbf{w}$ (weights for $\mathbf{x} \in \mathbf{X}$) \\
\hspace*{\algorithmicindent} \textbf{Output} Distance matrix $\mathbf{D}^{n \times n}$
\begin{algorithmic}[1]

\If {$T = \emptyset$}
	\State Update $d_{i,j} := d_{i,j} + 3 w_i w_j \:\: \forall \:\: i \neq j \:\: s.t. \:\: \mathbf{x}_{i}, \mathbf{x}_j \in \mathbf{X}$
	\State \Return $\mathbf{D}$
\Else
	\State Update $d_{i,j} := d_{i,j} + w_i w_j \:\: \forall \:\: i \neq j \:\: s.t. \:\: \mathbf{x}_{i}, \mathbf{x}_j \in \mathbf{X}$
	\If {chosen variable $y_T$ in $T$ is numeric}
		\State Determine subsets $\mathbf{X}_l = \{ \mathbf{x}_i \in \mathbf{X} | x_{i,y} \leq z_T \}$, $\mathbf{X}_r = \{ \mathbf{x}_j \in \mathbf{X} | x_{j,y} > z_T \}$
		\State Set weights $\mathbf{w}_l = \{ w_i \:|\: \mathbf{x}_i \in \mathbf{X}_l \} \cup \{ b_l w_i \:|\: x_{i,y}\: \text{is unknown} \}$
		\State Set weights $\mathbf{w}_r = \{ w_j \:|\: \mathbf{x}_j \in \mathbf{X}_r \} \cup \{ (1-b_l) w_j \:|\: x_{j,y}\: \text{is unknown} \}$
		\State Divide missing values s.t. $\mathbf{X}_l := \mathbf{X}_l \cup \{ \mathbf{x}_i \in \mathbf{X} \:|\: x_{i,y} \:\text{is unknown} \}$, $\mathbf{X}_r := \mathbf{X}_r \cup \{ \mathbf{x}_j \in \mathbf{X} \:|\: x_{j,y} \:\text{is unknown} \}$
	\Else
		\State Determine subsets $\mathbf{X}_l = \{ \mathbf{x}_i \in \mathbf{X} \:|\: x_{i,y} \in S_T \}$, $\mathbf{X}_r = \{ \mathbf{x}_j \in \mathbf{X} \:|\: x_{j,y} \notin S_T \land x_{j,y} \in \textsc{c}_T \}$
		\State Set weights $\mathbf{w}_l = \{ w_i \:|\: \mathbf{x}_i \in \mathbf{X}_l \} \cup \{ b_l w_i \:|\: x_{i,y} \:\text{is unknown}  \lor x_{i,y} \notin \textsc{c}_T \}$
		\State Set weights $\mathbf{w}_r = \{ w_j \:|\: \mathbf{x}_j \in \mathbf{X}_r \} \cup \{ (1-b_l) w_j \:|\: x_{j,y} \:\text{is unknown}  \lor x_{j,y} \notin \textsc{c}_T \}$
		 \State Divide missing values and unseen categories s.t. $\mathbf{X}_l = \mathbf{X}_l \cup \{ \mathbf{x}_i \in \mathbf{X} \:|\: x_{i,y} \:\text{is unknown} \lor x_{i,y} \notin \textsc{c}_T \}$, $\mathbf{X}_r = \mathbf{X}_r \cup \{ \mathbf{x}_j \in \mathbf{X} \:|\: x_{j,y} \:\text{is unknown}  \lor x_{j,y} \notin \textsc{c}_T \}$
	\EndIf
	\State \Return $\text{TraverseTree}(T_{\text{l}}, \mathbf{X}_l, \mathbf{D}, \mathbf{w}_l) + \text{TraverseTree}(T_{\text{r}}, \mathbf{X}_r, \mathbf{D}, \mathbf{w}_r) - \mathbf{D}$
\EndIf
\end{algorithmic}
\end{algorithm}

\begin{algorithm}[H]
\caption{TraverseExtTree}\label{TraverseExtTree}
\hspace*{\algorithmicindent} \textbf{Inputs} $T$ (node of an iTreeExtEnh), $\mathbf{X}$ (input data), $\mathbf{D}$ (current sum of separation depths) \\
\hspace*{\algorithmicindent} \textbf{Output} Distance matrix $\mathbf{D}^{n \times n}$
\begin{algorithmic}[1]

\If {$T = \emptyset$}
	\State Update $d_{i,j} := d_{i,j} + 3 \:\: \forall \:\: i \neq j \:\: s.t. \:\: \mathbf{x}_{i}, \mathbf{x}_j \in \mathbf{X}$
	\State \Return $\mathbf{D}$
\Else
	\State Update $d_{i,j} := d_{i,j} + 1 \:\: \forall \:\: i \neq j \:\: s.t. \:\: \mathbf{x}_{i}, \mathbf{x}_j \in \mathbf{X}$
	\State Initialize linear combination $\mathbf{y} := \mathbf{0}$ for each point in $\mathbf{X}$
	
	\For {each numeric variable $v \in \textsc{u}_T$}
		\State Update $y_i := y_i + 		
		\begin{cases}
      		z_v x_{i,v}, & x_{i,v} \:\: \text{is known} \\
      		r_v, & x_{i,v} \:\: \text{is unknown}
    	\end{cases}
    \forall x_{i,v} \in \mathbf{X}_{:,v}$
	\EndFor
	
	\For {each categorical variable $c \in \textsc{u}_T$}
		\State Update $y_i := y_i + 		
		\begin{cases}
      		s_{x_{i,v}}^c, & x_{i,v} \in s^c \\
      		r_c, & x_{i,v} \notin s^c \lor x_{i,v}  \: \text{is unknown}
    	\end{cases}
    \forall x_{i,v} \in \mathbf{X}_{:,v}$
	\EndFor
	
	\State Determine subsets $\mathbf{X}_l = \{ {\mathbf{x}_i \in \mathbf{X} \:|\: y_i \leq q_T} \}$, $\mathbf{X}_r = \{ {\mathbf{x}_j \in \mathbf{X} \:|\: y_j > q_T} \}$.
	
	\State \Return $\text{TraverseExtTree}(T_{\text{l}}, \mathbf{X}_l, \mathbf{D}) + \text{TraverseExtTree}(T_{\text{r}}, \mathbf{X}_r, \mathbf{D}) - \mathbf{D}$
\EndIf
\end{algorithmic}
\end{algorithm}

\section{Comparison to other distance metrics}

The metric proposed here (the implementation was made open source and freely available\footnote{\url{https://github.com/david-cortes/isotree}}) was compared against typical distance metrics (Euclidean, Mahalanobis, Cosine) in terms of their (Pearson) correlation under randomly-generated data with different properties, using the single-variable and the extended model with two variables at a time, both of them with (a) no sub-sampling, (b) full-depth trees, (c) 100 trees per model, (d) only-random splits.

The following comparisons take a randomly-generated matrix $\mathbf{X}$ composed of several column vectors $\mathbf{X} = 
\begin{bmatrix}
\mathbf{x}_1, & \mathbf{x}_2, & ..., & \mathbf{x}_n
\end{bmatrix}
$. Some of the values were later set randomly as missing for comparison purposes.

\begin{table}[H]
\caption {Independent variables with the same scale $\mathbf{x}_1, \mathbf{x}_2 \sim \text{Normal}(0, 1)$ - this is the kind of case in which Euclidean distance is the most approriate, and here it is equivalent to Mahalanobis due to the covariance matrix being an identity matrix.}
\begin{adjustbox}{max width=\textwidth}{\centering
\begin{tabular}{|r|c|c|c|c|}
 \hline
 \textbf{Metric} & Iso & IsoExt & Euc & Cos \\
 \hline
Iso &      & 0.944 & 0.951 & 0.622 \\
IsoExt & 0.944 &      & 0.968 & 0.62  \\
Euc & 0.951 & 0.968 &      & 0.628 \\
Cos & 0.622 & 0.62  & 0.628 &     \\
 \hline
\end{tabular}}\end{adjustbox}
\end{table}

\begin{table}[H]
\caption {Independent variables with different scale $\mathbf{x}_1 \sim \text{Normal}(0, 1)$, $\mathbf{x}_1 \sim \text{Normal}(0, 100)$ - here Euclidean distance will always weight the larger column heavier, but metrics such as Mahalanobis distance can easily overcome this difference.}
\begin{adjustbox}{max width=\textwidth}{\centering
\begin{tabular}{|r|c|c|c|c|c|}
 \hline
 \textbf{Metric} & Iso & IsoExt & Euc & Mah & Cos \\
 \hline
Iso &      & 0.944 & 0.671 & 0.95  & 0.382 \\
IsoExt & 0.944 &      & 0.697 & 0.971 & 0.378 \\
Euc & .671 & 0.697 &      & 0.697 & 0.542 \\
Mah & .95  & 0.971 & 0.697 &      & 0.361 \\
Cos & 0.382 & 0.378 & 0.542 & 0.361 &     \\
 \hline
\end{tabular}}\end{adjustbox}
\end{table}

\begin{table}[H]
\caption {Independent variables in the same scale, plus a non-linear transformation: $\mathbf{x}_1, \mathbf{x}_2 \sim \text{Normal}(0, 1)$, $\mathbf{x}_3 = \exp(\mathbf{x}_2)$ - intuitively, having the newly-added column which is just a deterministic transformation of an already-existing column does not add any extra information, so an ideal distance metric should be very similar to the simple Euclidean distance without the new column.}
\begin{adjustbox}{max width=\textwidth}{\centering
\begin{tabular}{|r|c|c|c|c|c|c|c|c|}
 \hline
 \textbf{Metric} & Iso & IsoExt & Euc & Mah & Cos & Euc (no $\mathbf{x}_3$) & Mah (no $\mathbf{x}_3$) & Cos (no $\mathbf{x}_3$) \\
 \hline
Iso &      & 0.962 & 0.657 & 0.768 & 0.605 & 0.924 & 0.924 & 0.551 \\
IsoExt & .962 &      & 0.72  & 0.832 & 0.619 & 0.929 & 0.93  & 0.522 \\
Euc & .657 & 0.72  &      & 0.916 & 0.177 & 0.563 & 0.562 & 0.234 \\
Mah & .768 & 0.832 & 0.916 &      & 0.454 & 0.761 & 0.76  & 0.383 \\
Cos & 0.605 & 0.619 & 0.177 & 0.454 &      & 0.747 & 0.747 & 0.756 \\
Euc  (no $\mathbf{x}_3$) & 0.924 & 0.929 & 0.563 & 0.761 & 0.747 &      &      & 0.628 \\
Mah  (no $\mathbf{x}_3$) & 0.924 & 0.93  & 0.562 & 0.76  & 0.747 &      &      & 0.628 \\
Cos  (no $\mathbf{x}_3$) & 0.551 & 0.522 & 0.234 & 0.383 & 0.756 & 0.628 & 0.628 &     \\
 \hline
\end{tabular}}\end{adjustbox}
\end{table}

\begin{table}[H]
\caption {Non-independent variables $\mathbf{x}_1, \mathbf{x}_2, \mathbf{x}_3, \mathbf{x}_4, \mathbf{x}_5 \sim \text{Normal}(\mu, \Sigma)$, $\mu = \protect\begin{bmatrix} 0.619,& 2.149,& 0.083,& 0.113,& 3.66  \protect\end{bmatrix}$, $\Sigma = 
\protect\begin{bmatrix}
6.17 &  1.87 & -2.82 & -1.35 & -1.48  \protect\\
1.87 &  3.01 & -1.03 & -0.84 &  1.56  \protect\\
-2.82 & -1.03 &  3.94 & -0.8  & -0.73  \protect\\
-1.35 & -0.84 & -0.8  &  1.67 &  0.59  \protect\\
-1.48 &  1.56 & -0.73 &  0.59 &  2.77
\protect\end{bmatrix}
$ (all of these distribution parameters were randomly-generated and do not represent anything meaningful) - this is the kind of scenario in which Mahalanobis distance is the most appropriate  as variables are only related by their linear correlation, under a single unimodal distribution from which all of them are drawn. Additionally, a random 15\% of the values was set as missing, and in the case of Euclidean, Mahalanobis, and Cosine distance, was imputed with the column mean.}
\begin{adjustbox}{max width=\textwidth}{\centering
\begin{tabular}{|r|c|c|c|c|c|c|c|c|c|c|}
 \hline
 \textbf{Metric} & Iso & IsoExt & Euc & Mah & Cos & Iso (15\% NA) & IsoExt (15\% NA) & Euc (15\% NA) & Mah (15\% NA) & Cos (15\% NA) \\
 \hline
Iso &      & 0.96 & 0.94 & 0.74 & 0.7  & 0.63 & 0.86 & 0.85 & 0.78 & 0.63 \\
IsoExt & 0.96 &      & 0.94 & 0.76 & 0.67 & 0.6  & 0.87 & 0.86 & 0.79 & 0.6  \\
Euc & 0.94 & 0.94 &      & 0.75 & 0.74 & 0.6  & 0.85 & 0.9  & 0.78 & 0.66 \\
Mah & 0.74 & 0.76 & 0.75 &      & 0.56 & 0.48 & 0.69 & 0.68 & 0.72 & 0.5  \\
Cos & 0.7  & 0.67 & 0.74 & 0.56 &      & 0.47 & 0.58 & 0.63 & 0.51 & 0.87 \\
Iso (15\% NA) & 0.63 & 0.6  & 0.6  & 0.48 & 0.47 &      & 0.5  & 0.52 & 0.57 & 0.43 \\
IsoExt (15\% NA) & .86 & 0.87 & 0.85 & 0.69 & 0.58 & 0.5  &      & 0.94 & 0.79 & 0.66 \\
Euc (15\% NA) & 0.85 & 0.86 & 0.9  & 0.68 & 0.63 & 0.52 & 0.94 &      & 0.79 & 0.7  \\
Mah (15\% NA) & 0.78 & 0.79 & 0.78 & 0.72 & 0.51 & 0.57 & 0.79 & 0.79 &      & 0.55 \\
Cos (15\% NA) & 0.63 & 0.6  & 0.66 & 0.5  & 0.87 & 0.43 & 0.66 & 0.7  & 0.55 &     \\
 \hline
\end{tabular}}\end{adjustbox}
\end{table}

\begin{table}[H]
\caption {Gaussian mixture with non-independent variables and equal probability for each group - $\mathbf{x}_1^a, \mathbf{x}_2^a \sim \text{Normal}(\protect\begin{bmatrix} -1, & -1 \protect\end{bmatrix},
\protect\begin{bmatrix}
0.1  &  -0.2  \protect\\
-0.2 &  0.5
\protect\end{bmatrix},
)$, $\mathbf{x}_1^b, \mathbf{x}_2^b \sim \text{Normal}(\protect\begin{bmatrix} 0.25, & 0.25 \protect\end{bmatrix},
\protect\begin{bmatrix}
0.1  &  0.2  \protect\\
0.2 &  0.5
\protect\end{bmatrix},
)$. Here an ideal metric should make points within groups closer  than points between groups, and the metric should take into account the internal correlations within each group more than the mixed correlations (this is shown in the table at the right). Best reference here is Euclidean distance, but it still doesn't account for relationships between variables. Since the covariance matrix is the same but with oposite signs at the non-diagonal entries, under an ideal metric, the average distance between points within group $a$ should be similar to the average distance between points within group $b$.}
\begin{adjustbox}{max width=\textwidth}{\centering
\begin{tabular}{|r|c|c|c|c|c|}
 \hline
 \textbf{Metric} & Iso & IsoExt & Euc & Mah & Cos \\
 \hline
Iso &    & 0.97 & 0.95 & 0.89 & 0.69 \\
IsoExt & 0.97 &     & 0.96 & 0.87 & 0.76 \\
Euc & 0.95 & 0.96 &     & 0.9  & 0.71 \\
Mah & 0.89 & 0.87 & 0.9  &     & 0.55 \\
Cos & 0.69 & 0.76 & 0.71 & 0.55 &    \\
 \hline
\end{tabular}}\end{adjustbox}
\begin{adjustbox}{max width=\textwidth}{\centering
\begin{tabular}{|r|c|c|c|c|c|}
 \hline
 \textbf{Metric} & Iso & IsoExt & Euc & Mah & Cos \\
 \hline
$\bar{d}(\mathbf{x}^a$, $\mathbf{x}^a)$ & 0.3 & 0.27 & 0.94 & 1.51 & 0.2 \\
$\bar{d}(\mathbf{x}^b$, $\mathbf{x}^b)$ & 0.28  & 0.27 & 0.88 & 0.92 & 0.84 \\
$\bar{d}(\mathbf{x}^a$, $\mathbf{x}^b)$ & 0.54 & 0.58 & 1.96 & 2.26 & 1.35 \\
 \hline
\end{tabular}}\end{adjustbox}
\label{table:mixture}
\end{table}

\begin{figure}[tph!]
\centerline{\includegraphics[totalheight=5cm]{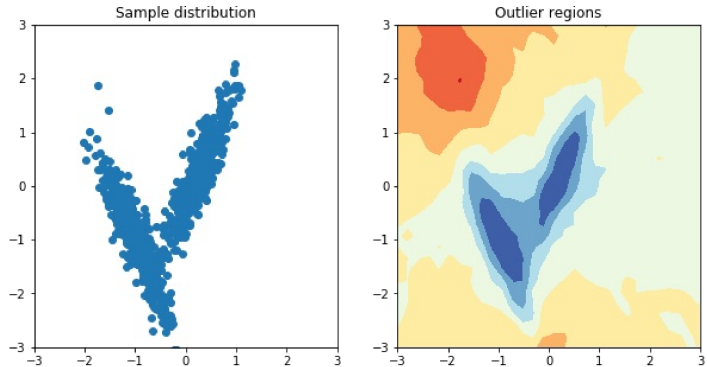}}
    \caption{Sample points from mixture used in example \ref{table:mixture} (outlier regions are from extended model).}
    \label{fig:verticalcell}
\end{figure}

In these examples, the most suitable metric under each specific situation presented the highest correlation with the distance metric proposed here, even though the reverse was not always the case. The extended model shows a slight edge in most cases, which becomes a rather large edge in the case of missing values as it was able to maintain a higher correlation against the same distance obtained when the values are not missing. Both models were able to produce comparable within-group distances in a mirrored Gaussian mixture, which a distance such as Mahalanobis that takes the mixed covariance matrix cannot do.

As a more realistic comparison point, the metric proposed here for the extended model was also compared against Gower distance (calculated using the R package "cluster" with its default parameters - see \cite{cluster}) under the hypothyroid dataset\footnote{https://archive.ics.uci.edu/ml/datasets/Thyroid+Disease}, which contains a mixture of numeric, boolean, and categorical variables, with missing values in several of them and non-normally-distributed numeric variables, this time with limited-depth trees and some non-random splits as in \cite{sciforest} - the (Pearson) correlation between these metrics stood at $0.731$. Unfortunately, for such kind of data, it's very difficult to make a detailed comparison and/or determine which one produces the most desirable output, so the comparison was stopped at that.

\section{Conclusions}

This work introduced a metric distance between points in an arbitrary feature space which is obtained with the use of Isolation Forest models and is based on a sample from the data-generating distribution. Compared to more typical metrics such as Euclidean or Mahalanobis distance, this metric was shown to be more robust against different possible relationships between variables, to produce more desirable relative distances under mixed distributions, and to have other desirable properties such as being limited in range and having a threshold value that can be used to determine if two points are more similar than dissimilar. Some simple extensions to the Isolation Forest algorithm were proposed to allow calculations with missing values and categorical variables, which in the case of missing values was shown to provide highly-correlated results with the non-missing-data distance, and in the case of mixed numeric and categorical variables, was shown to correlate highly with Gower distance, while still being able to account for relationships between numeric and categorical variables.

\renewcommand{\abstractname}{Acknowledgements}
\begin{abstract}
The derivation and formulas of the expected isolation and separation depths in randomly-generated trees shown here were thanks to users Misha Lavrov and antkam in the StackExchange website for mathematics, who kindly provided detailed explanations of the logic and the formulas when these questions were asked there.
\end{abstract}

\bibliographystyle{plain}
\bibliography{iso}

\end{document}